\pdfoutput=1
\documentclass{article}
\usepackage{spconf,amsmath,graphicx,booktabs,microtype,multirow,longtable,array}
\usepackage[hidelinks]{hyperref}

\title{Coupled but Late: Turn-Taking Between\\ Full-Duplex Speech Models in Unscripted Dialogue}

\name{Lichen Zhu, Yueqian Lin, Yiheng Wang, Hai ``Helen'' Li, Yiran Chen\thanks{Preprint. Submitted to ICASSP 2027. The appendices carry the statistical notes behind every reported number.}}
\address{Duke University, Durham, NC, USA}

\begin{document}
\maketitle

\begin{abstract}
Full-duplex speech models are trained to converse with a person, but they
are increasingly made to converse with each other, in self-play data
generation, agent societies, and model-based evaluation. In that loop no
human absorbs a timing error: each model's turn-taking is the other's
input. We ask what timing the loop settles into. Two PersonaPlex-7B
instances exchange audio tokens on a shared clock in unscripted
conversation, and one floor-transfer rule is applied to them and to
Switchboard. Their timing is coupled: re-pairing speakers across
conversations destroys it. But the floor changes hands late, at a median
of 400--560\,ms against 137\,ms for humans, and the last 120\,ms of the
partner's turn, where human projection places a tenth of its transfers,
holds 1\% of theirs. Delaying one direction of the channel shifts the response one-for-one and
leaves the run-up to it empty, consistent with a reactive wait after the
perceived end rather than the turn-end projection human timing requires.
\end{abstract}


\section{Introduction}
\label{sec:intro}

A full-duplex speech model listens while it speaks and decides for itself
when to take the floor~\cite{defossez2024moshi,roy2026personaplex}. It is
trained and evaluated as one side of a conversation whose other side is a
person or a recording standing in for one. Increasingly, though, two such
models are placed in a loop with no person in it: to generate self-play
dialogue data~\cite{wang2026agentic,kim2026duplexgen}, to populate spoken
agent societies~\cite{park2023}, or to act as an automated examiner for
another model~\cite{lin2026fdbench2}. In that loop nobody absorbs a timing error: whatever one model does at
the floor is the input on which the other decides, and what timing the
loop settles into has no answer in single-model evaluation.

Benchmarks score one model against fixed stimuli or an
examiner~\cite{lin2025fdbench,lin2026fdbench15,arora2025talkingturns,lin2026fdbench2},
and the model--model dyads built so far serve other questions: SyncLLM
compares two agents' gap and overlap statistics with human continuations
under communication delay~\cite{veluri2024syncllm}, dGSLM generates both
channels from one model and tabulates them against
Fisher~\cite{nguyen2023dgslm}, Riera et al.~\cite{riera2026} route audio
tokens between two Moshi instances, as we do, to study representational
synchronization, and Agentic-DuplexGen has two duplex models perform a
written script and reports their floor-transfer
offsets~\cite{wang2026agentic}. None measures floor timing for freely generated dyads under controls, or
asks whether a difference from human timing belongs to the models or to
the channel between them. On the policy side the question has
a sharper form. Classical dialogue systems end-point on a silence
threshold, so they can only react to a turn that has already
ended~\cite{skantze2021review}; humans project turn ends and start 100 to
200\,ms after them~\cite{levinson2015}, and projection models such
as VAP are built to do the same~\cite{ekstedt2022vap}. Which of the two a
duplex model implements when it talks to another duplex model is not known.

We answer it for PersonaPlex-7B. One floor-transfer rule, applied by the
same code to both agents and to Switchboard word alignments, gives the
timing; three agent-side segmentations bound its dependence on speech
boundaries; re-pairing speakers and silencing the partner test what the
timing depends on; and a latency adjustment and a one-sided delay sweep
test whether the channel explains it. The timing is coupled, it is late,
and the lateness behaves like a reactive policy rather than like the
channel's delay. A secondary intervention shows that the opening register
the models bring from training yields to a sentence placed in one agent's
speech.

\begin{figure*}[t]
  \centering
  \includegraphics{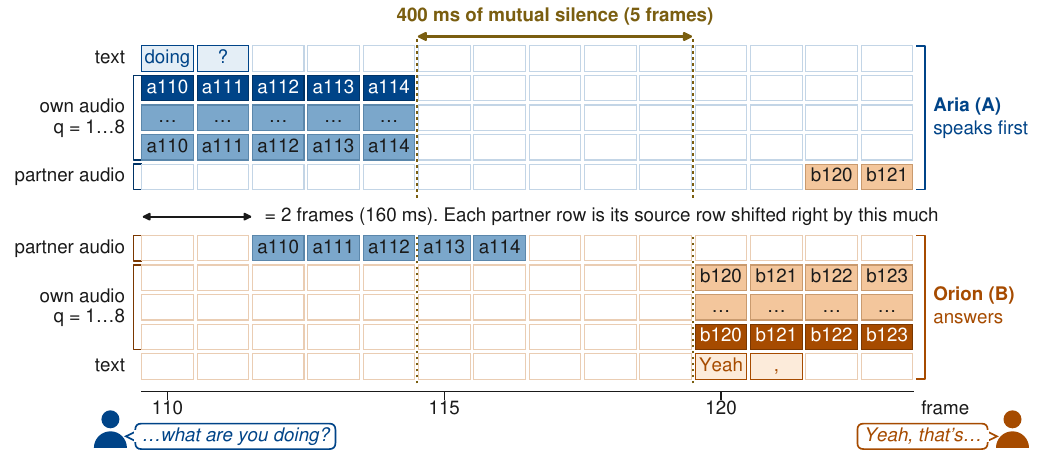}
  \caption{Direct audio-token exchange. Colors identify token origin:
  A (blue) and B (orange). Each speaker outputs one text stream and eight
  audio codebooks (two drawn). Partner tokens arrive after 160\,ms
  (coarse, illustrated) or 240\,ms (fine). The example gap is 400\,ms;
  the baseline mode is 480\,ms.}
  \label{fig:apparatus}
\end{figure*}

\section{Setup and Measurement}
\label{sec:setup}

\textbf{Conversations.} Two PersonaPlex-7B agents
(\texttt{nvidia/}\allowbreak\texttt{personaplex-7b-v1}), one per GPU,
exchange Mimi audio tokens~\cite{defossez2024moshi} at 12.5\,Hz
(Fig.~\ref{fig:apparatus}). At each 80\,ms step an agent receives the
tokens its partner emitted at the previous step; nothing on the path is
audio, so no recognizer, synthesizer, or re-encoding adds its own timing.
A partner's frame becomes readable after 160\,ms on the coarse codebook and
240\,ms on the fine ones, from one frame of buffering plus the model's
codebook schedule; an agent can still start before its partner stops. The baseline is 36 seeded 90\,s conversations in five scenes that put both
speakers in one place, such as an apartment laundry room. Personas and
scenes enter as system prompts, temperatures are 0.8 for audio and 0.7 for
text, and every matched condition keeps personas, scenes, voices, and seeds
fixed; audio is decoded afterwards for measurement only.

\begin{table*}[t]
  \centering\small
  \caption{Floor-transfer timing under one extraction rule, grouped by the
  claim each block supports; bold marks the cells the claim rests on.
  Offsets and $W_1$ in ms; overlap ($\mathrm{FTO}<-10$\,ms, any depth),
  $[-120,0)$ (the partner's last 120\,ms) and $\pm120$ (the near-zero band)
  are fractions of transfers. Under a B$\to$A delay of $D$ frames the delayed
  side's mode is written as 480\,ms plus the delay. Brackets are 95\%
  intervals: conversation bootstrap, the central range of 2{,}000
  re-pairings, and the sampling null.}
  \label{tab:big}
  \setlength{\tabcolsep}{3pt}
  \begin{tabular}{@{}llrrrrrrr@{}}
    \toprule
    & condition & $n$ & median & mode & overlap & $[-120,0)$ & $\pm120$ & $W_1$\\
    \midrule
    \multirow{3}{*}{\shortstack[l]{human\\reference}}
      & Switchboard, full calls & 193{,}115 & $+137$ & 160 & 0.32 & 0.10 & 0.25 & --\\
      & first 90\,s of each call & 43{,}890 & $+163$ & 160 & 0.28 & 0.10 & 0.25 & --\\
      & sampling null, $n=747$ & & & & & & & 50 $[18,123]$\\
    \midrule
    \multirow{3}{*}{late}
      & energy (primary) & 747 & $\mathbf{+480}$ $[400,480]$ & 480 & 0.18 $[.15,.20]$ & \textbf{0.01} & 0.03 $[.01,.05]$ & 228 $[208,257]$\\
      & text words & 805 & $\mathbf{+560}$ & 640 & 0.14 & \textbf{0.02} & 0.09 & 330\\
      & aligned words & 736 & $\mathbf{+400}$ & 400 & 0.18 & \textbf{0.02} & 0.04 & 226\\
    \midrule
    \multirow{3}{*}{coupled}
      & partner re-paired & -- & $+80$ $[0,160]$ & -- & 0.46 $[.42,.49]$ & -- & -- & \textbf{520} $[471,570]$\\
      & partner silenced, 12 cells & \textbf{26}/273 & -- & -- & -- & -- & -- & --\\
      & recorded human partner & 384 & $+240$ & \textbf{320} & 0.22 & 0.03 & 0.16 & 103\\
    \midrule
    \multirow{5}{*}{\shortstack[l]{not the\\channel}}
      & energy, $-240$\,ms & 747 & $\mathbf{+240}$ & 240 & 0.21 & 0.02 & 0.12 & 173\\
      & B$\to$A delay, $D=3$ & 411 & $+640$ & \textbf{480}+240 & 0.28 & 0.01 & 0.05 & 321\\
      & $D=8$ & 410 & $+1040$ & \textbf{480}+640 & 0.32 & 0.02 & 0.07 & 495\\
      & $D=10$ & 395 & $+1040$ & \textbf{480}+800 & 0.38 & 0.02 & 0.05 & 561\\
      & $D=15$ & 361 & $+720$ & \textbf{480}+1200 & 0.36 & 0.02 & 0.06 & 722\\
    \bottomrule
  \end{tabular}
\end{table*}

\textbf{Floor-transfer offset.} Speech spans separated by less than
180\,ms merge into inter-pausal units (IPUs), and following Heldner and
Edlund~\cite{heldner2010} a floor transfer is the interval from the end of
the current speaker's IPU to the start of the next speaker's,
\begin{equation}
  \mathrm{FTO}=t_{\mathrm{start,next}}-t_{\mathrm{end,current}},
\end{equation}
positive for a gap and negative for an overlap. An IPU wholly inside the other speaker's does not transfer the floor;
identical-extent IPUs are resolved by who holds the floor, so counts do not
depend on speaker order. No lexical backchannel filter is applied~\cite{heldner2011vsu}; under the
shared rule, nested IPUs are 14\% of different-speaker adjacencies for
humans against 14--16\% for the agents, and sub-second IPUs 43\% against
45\% under the primary instrument (50\% and 84\% under the word
segmentations, which cut at word boundaries).
Offsets are clipped at $\pm2$\,s. We report the median, the overlap
fraction ($\mathrm{FTO}<-10$\,ms, a margin that excludes exact zeros on
the 80\,ms grid), the share within $\pm120$\,ms of zero, and the
Wasserstein-1 distance $W_1$ between the clipped distributions, in
milliseconds. The $\pm120$\,ms window, the \emph{near-zero band}, follows
Heldner's detection thresholds for gaps and overlaps~\cite{heldner2011}.

\textbf{Reference and segmentation.} The same extraction function runs on
the released MS-State word alignments for
Switchboard~\cite{godfrey1992,deshmukh1998}: 193{,}115 transfers from
2{,}407 calls with at least 20 each. Its median of 137\,ms sits below the 168\,ms of Roberts et
al.~\cite{roberts2015} because IPU-level segmentation with backchannels
retained counts shorter intervals. Switchboard is telephone speech between strangers while the agents share
a scene; the two settings differ, and the recorded-partner arm of
Sec.~\ref{sec:recorded} gives the co-present comparison. Agent speech has no aligner, so we segment it three ways: decoded-audio
energy above 0.01 RMS, the primary instrument, which applies unchanged to
live and recorded partners and carries through every control; word extents
from the text stream; and words aligned to the voiced runs that contain
them. The rule is shared, the boundary estimators are not, and the spread
across the three is a sensitivity result.

\textbf{Inference.} Bootstrap intervals resample conversations with the
human reference fixed; paired intervals invert the sign-flip test. The
reporting cutoff is $0.05/45$, a conservative divisor kept to avoid
post-hoc relaxation; the baseline timing contrasts, the re-pairing control,
and the whole-conversation opener contrasts pass it, the rest is nominal.
The appendices carry the statistical notes with the enumerated tests and
the clip and grid sensitivity checks.

\begin{figure}[t]
  \centering
  \includegraphics{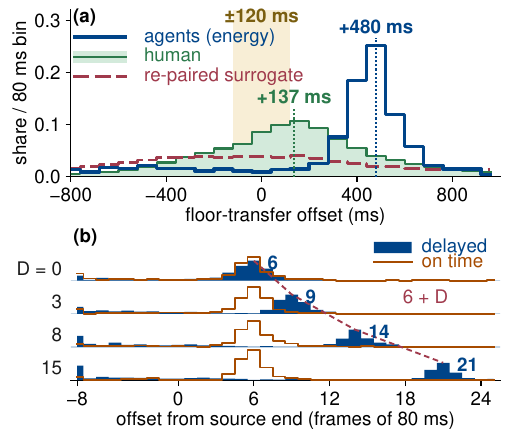}
  \caption{(a) Energy-based agent offsets, the full-call human reference,
  and the re-paired surrogate; 80\,ms bins, shading marks $\pm120$\,ms.
  (b) Delaying B$\to$A by $D$ frames (four of the five levels) moves A's peak to $6+D$ (filled) while B's stays at six (line); rows share one scale, one frame is 80\,ms.}
  \label{fig:fto}
\end{figure}

\section{Results}
\label{sec:results}

\subsection{The floor changes hands late}
\label{sec:late}

Table~\ref{tab:big} and Fig.~\ref{fig:fto}a compare the baseline dyads
with full-call Switchboard. Under every segmentation the agents' median offset is three to four times
the human one, and the near-zero band, which holds a quarter of human
transfers, holds 3 to 9\% of theirs. The
distribution is not merely wide: 44\% of all transfers fall at 400 or
480\,ms, a sharp peak with no counterpart in the human histogram.
Fig.~\ref{fig:case}a shows one opening: after a simultaneous start, the
agents hand over at 320 to 640\,ms and once at 1360\,ms, the Switchboard callers at 9 and 45\,ms, once
in overlap, and twice after longer gaps. Measurement does not manufacture this. The three segmentations move the
median by up to 160\,ms and leave both deficits in place; the
first-90-second reference, which matches conversation phase, brings the
human side closer and leaves them too; and the observed $W_1$ is several times the sampling null of
Table~\ref{tab:big}, with every one of the 36 conversations above its own
size-matched null (sign test, $p=2.9\times10^{-11}$). Shifting the human reference to the agents' median isolates shape from
location: the agents overlap \emph{more} than a merely delayed human
would, so their lower overlap is the same lateness seen again, not a second
deficit, and the near-zero
deficit survives the shift under two of the three segmentations.

\subsection{The timing is coupled}
\label{sec:coupled}

A shared clock alone does not make two speakers a conversation, so the
first control asks whether the timing belongs to the pair. Pairing agent A
of conversation $i$ with agent B of conversation $j\neq i$ keeps each
speaker's activity track and the clock and removes only the original
partner; two monologues on a common clock would be unaffected.
Instead (Table~\ref{tab:big}) the re-paired median collapses toward zero,
overlap more than doubles, the interquartile range widens from 160 to
1360\,ms, and the distance to the human reference grows; the real dyads sit
outside all 2{,}000 re-pairings on every statistic. Same-scene re-pairings reproduce the free surrogate, so the structure
belongs to the original pairing, not to the clock or the scene. The second control replaces the partner's audio with encoded silence in
twelve matched cells: transfers fall from 273 to 26 and words per agent
from 134 to 29; the agents say their opening line and stop. Partner audio
sustains speaking at all, a production result rather than a timing one,
but it rules out agents talking past each other on a schedule.

\subsection{Lateness is a policy, not the channel}
\label{sec:policy}

The channel adds up to 240\,ms of delay, so the obvious explanation is the
link. Two observations rule it out as the
whole story. Subtracting the full 240\,ms from every transfer (Table~\ref{tab:big},
``$-$240'') leaves a median of 240\,ms, still above both human references,
and the near-zero band at half the human share; a constant cannot fix a
shape. And the last 120\,ms of the partner's turn, where projection shows, is
nearly empty: humans place a tenth of their transfers there, the agents
one percent, two with the link delay removed. The agents do overlap, on
18\% of transfers, but almost all of that mass lies deeper inside the partner's turn (beyond
120\,ms, where the median is $-560$\,ms against $-311$\,ms): interruption,
not projection~\cite{levinson2015}.

The delay sweep asks what the wait is anchored to: B's tokens reach A
$D\in\{3,8,10,15\}$ frames late while A's reach B on time, 36 conversations
per level. A's pooled response mode moves by exactly $D$ at every level while B's
stays at six frames (Fig.~\ref{fig:fto}b; Table~\ref{tab:big}); per conversation the modal wait beyond the delay has a median of four and a half to six frames and lies within one frame of
six in 18 to 28 of the 35 or 36
conversations at each level, with ties between modal frames resolved to the
earlier one. Measured from the moment B's end becomes visible to A, the pooled wait is
unchanged: it rides on the partner's audio, and a partner that appears up to
1.2\,s slower does not shorten the mode. Its shape says what kind of wait it is. Nine in ten of the agents'
positive offsets begin more than 320\,ms after the partner's end, against
four in ten for humans, and the approach to that end holds one to two percent at
every level. A late mode with the run-up to it unoccupied is consistent with a
silence-threshold policy~\cite{skantze2021review} and not with projection,
which places part of its onsets in that run-up, delayed or not. After
the 240\,ms link, the model waits roughly one further link delay of silence
before it speaks. Chang et al.~\cite{chang2026inertia} measure a 0.6\,s generate-to-perceive
lag in PersonaPlex during interruptions and call it state inertia; the
wait here is consistent with it, though floor timing alone cannot identify
the mechanism.

\subsection{A recorded partner does not elicit it}
\label{sec:recorded}

Replacing the live partner with recorded human speech (72 runs over 24
windows from two AMI headset recordings~\cite{carletta2005ami},
level-matched and delivered one frame late for parity)
removes the six-frame peak: that frame holds a third as many transfers as
in live dyads, the median halves (Table~\ref{tab:big}), and the band share (0.16) sits inside the range of a mismatched-window surrogate (0.11 to 0.19). A recording never yields the floor to the agent, and the agents mostly do
not take it;
their speech share halves. The 480\,ms wait is therefore not a fixed
latency applied to any input but what the model does when a partner yields.
Replacing the partner also changes content and acoustics, so this arm shows
condition dependence, not a closed-loop effect.

\begin{figure}[t]
  \centering
  \includegraphics{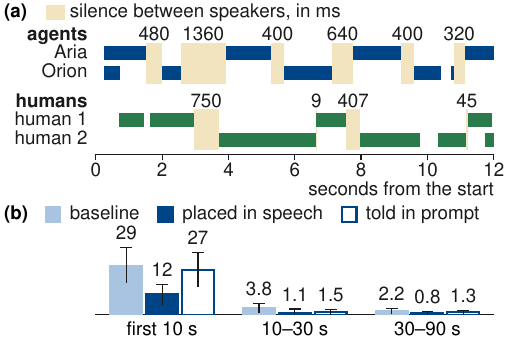}
  \caption{(a) The first 12\,s of conversation~1 (the pair of
  Fig.~\ref{fig:apparatus}) and of Switchboard call 2009, the first call
  with five transfers in that span. (b) Telephone-register phrases per thousand words by window, means over the conversations with words in the window; thin lines are 95\% conversation-bootstrap intervals.}
  \label{fig:case}
\end{figure}

\section{The Opening Register}
\label{sec:ritual}

The loop propagates more than timing. Placed in a laundry room, the models
open in telephone or customer-service language: conversation~1 begins
``Hi, this is Aria.'' ``Hi, Orion.'' ``Hi, how are you?''
(Fig.~\ref{fig:case}a). A 21-pattern lexicon written after inspecting the
baseline counts 29 such phrases per thousand
words in the first ten seconds
and 2 to 4 afterwards (Fig.~\ref{fig:case}b), consistent with the
checkpoint's customer-service training data~\cite{roy2026personaplex}.
Placing one of five scene-appropriate openers in one agent's initial
speech, with its timing left to the agent, cuts the opening rate from 29 to
12 ($-16.9$ $[-27.7,-6.0]$, nominal) and the whole-conversation rate from
7.6 to 2.2 ($-5.4$ $[-8.8,-2.2]$; $-5.3$ with the forced tokens excluded).
The effect sits with the agent that speaks the opener ($-7.4$
$[-11.2,-3.7]$); its partner moves the same way without reaching the cutoff
($-3.5$ $[-7.9,+0.8]$). The same sentence given as a system-prompt
instruction is followed in only 8 of 36 runs and leaves the opening where it
was (26.7), for $-2.6$ $[-5.8,+0.6]$ over the conversation: placed in the
speech, the sentence reaches the register; told in the prompt, it does not
reach the opening.
Instruction following in duplex models is a benchmark target in its own
right~\cite{tang2026instructfd}.

\section{Conclusion}
\label{sec:concl}

Two full-duplex models left to talk to each other are coupled but late:
the floor changes hands three to four times later than between humans, at a wait anchored to the end of the partner's audio, and the last 120\,ms of the
partner's turn, where human projection places a tenth of its transfers,
goes almost unused. Coupling is therefore no evidence of human-like timing,
and dialogue generated by such a pair, or an examiner built from such a model, inherits a reactive endpointer's floor. The result is for one checkpoint, with related
lateness reported for its model
family~\cite{chang2026inertia,ohashi2026alignment}; the protocol applies
unchanged to any duplex model. Code and measurements will be released; the appendices carry the
statistical notes.

\section{Compliance with Ethical Standards}
\label{sec:ethics}

This study used the publicly released MS-State word alignments for
Switchboard and the AMI Meeting Corpus recordings, under their respective
licenses. It involved no new human participants and required no new ethics
approval.

\bibliographystyle{IEEEbib}
\bibliography{refs}

\clearpage
\onecolumn
\appendix
\setcounter{table}{0}
\renewcommand{\thetable}{A\arabic{table}}
\section*{Appendices: statistical notes}
These appendices are the audit trail behind the paper: the inferential
statements behind the reporting cutoff, the intervals and sensitivity
checks, and every condition in full, including rows the paper omits for
space. Table and figure numbers without a prefix refer to the paper.

\section{Scope and reporting threshold}
The multiplicity inventory enumerates 31 tests, reproduced in
Table~\ref{tab:inventory}. The paper's cutoff is $\alpha_*=0.05/45\approx
0.0011$, a conservative divisor retained from the first analysis pass
rather than relaxed after inspecting the results. Contrasts below the
cutoff are the baseline timing contrasts (rows 1--9), the re-pairing controls (10--17), the silence control (22), the whole-conversation
forced-opener contrasts (26, 28, 29), and the per-conversation $W_1$ sign
test (31). Tables~\ref{tab:rows} and \ref{tab:opener} carry the nominal-level
contrasts the paper also reports. All 95\% intervals are marginal, not
simultaneous.

\section{Procedures}
\noindent\textbf{Conversation bootstrap (CB).}
The resampling unit is the agent conversation; all transfers of a sampled
conversation are pooled before each statistic is recomputed. The human
reference is fixed, so intervals quantify agent-side uncertainty only.
Table~1's primary row and every tail probability in the inventory
(Table~\ref{tab:inventory}, rows 1--9, all three segmentations) use 10{,}000
draws, so the reporting floor $2/(B+1)$ is 0.0002; the intervals of the
other rows in Table~\ref{tab:rows} use 1{,}000 draws (300 for $W_1$) and
carry no tail probability. Two-sided tail
probabilities against the fixed human statistic $T_H$ are
$\min\{1,\,2\min((1+\#\{T_b\le T_H\})/(B+1),\,(1+\#\{T_b\ge T_H\})/(B+1))\}$,
with reporting resolution $2/(B+1)$.

\noindent\textbf{Re-pairing surrogate (DP).}
Each of 2{,}000 derangements pairs one speaker's complete activity track
with a partner track from a different conversation, keeping the shared
clock. The two-sided probability is $(1+c)/(B+1)$, where $c$ counts
surrogate statistics at least as far from the surrogate median as the
observed one; the floor is $1/2001$. Table~1's re-paired row gives the
central 95\% of the surrogate statistics, not a confidence interval for the
observed value. Restricting derangements to pairs staged in the same scene
reproduces the free surrogate to within 0.005 of overlap and 0\,ms of
median.

\noindent\textbf{Paired sign-flip contrasts (SF).}
Within each conversation and window the rate is $1000\times$markers/words
pooled over both agents; the effect is the mean of paired per-conversation
differences over pairs with at least one word in both arms. Two-sided
probabilities use 20{,}000 random sign flips with the $(1+c)/(B+1)$
correction; intervals invert the same procedure and include zero whenever
its null is accepted. The procedure assumes sign-exchangeability of the
paired differences under the null. Excluding the forced opener removes its
tokens from numerator and denominator alike.

\noindent\textbf{Exact sign tests (EB).}
The silence control uses the 12 matched persona-pair and scene cells; all
12 have more transfers with partner audio ($p=2/2^{12}=4.9\times10^{-4}$).
The $W_1$ check uses the 36 baseline conversations; each exceeds the median
of its own size-matched clustered null ($p=2/2^{36}=2.9\times10^{-11}$).

\section{Every condition, with intervals}
Table~\ref{tab:rows} extends Table~1 with intervals for every baseline
row and the on-time direction of each delay level. The
contract is the same throughout: IPUs from spans separated by less than
180\,ms; a floor transfer from the current speaker's IPU end to the next
speaker's IPU start; nested IPUs excluded; identical-extent IPUs resolved
by the floor holder; offsets clipped at $\pm2$\,s.

\begin{table}[ht]
\centering\small
\caption{Every condition under the energy instrument unless noted. Median,
mode, and $W_1$ in ms; brackets are 95\% conversation-bootstrap intervals.
The two delay columns per level are the delayed direction (B$\to$A) and
the on-time direction (A$\to$B) of the same conversations.}
\label{tab:rows}
\setlength{\tabcolsep}{3pt}
\begin{tabular}{@{}lrrrrrrr@{}}
\toprule
condition & $n$ & median & mode & overlap & $[-120,0)$ & $\pm120$ & $W_1$\\
\midrule
energy (primary) & 747 & $+480$ $[400,480]$ & 480 & 0.18 $[.15,.20]$ & 0.011 & 0.03 $[.01,.05]$ & 228 $[208,257]$\\
text words & 805 & $+560$ $[560,640]$ & 640 & 0.14 $[.12,.16]$ & 0.024 & 0.09 $[.07,.11]$ & 330 $[293,367]$\\
aligned words & 736 & $+400$ $[400,480]$ & 400 & 0.18 $[.15,.21]$ & 0.020 & 0.04 $[.02,.06]$ & 226 $[204,260]$\\
energy, $-$240\,ms & 747 & $+240$ $[160,240]$ & 240 & 0.21 $[.17,.25]$ & 0.024 & 0.12 $[.08,.16]$ & 173 $[150,202]$\\
\midrule
$D=0$, B$\to$A & 377 & $+400$ & 480 & 0.17 & 0.013 & 0.04 & 227\\
$D=0$, A$\to$B & 370 & $+480$ & 480 & 0.18 & 0.008 & 0.02 & 229\\
$D=3$, delayed & 411 & $+640$ & 720 & 0.28 & 0.012 & 0.05 & 321\\
$D=3$, on time & 408 & $+480$ & 480 & 0.11 & 0.015 & 0.04 & 282\\
$D=8$, delayed & 410 & $+1040$ & 1120 & 0.32 & 0.017 & 0.07 & 495\\
$D=8$, on time & 398 & $+480$ & 480 & 0.10 & 0.005 & 0.02 & 289\\
$D=10$, delayed & 395 & $+1040$ & 1280 & 0.38 & 0.015 & 0.05 & 561\\
$D=10$, on time & 385 & $+480$ & 480 & 0.11 & 0.005 & 0.02 & 284\\
$D=15$, delayed & 361 & $+720$ & 1680 & 0.36 & 0.017 & 0.06 & 722\\
$D=15$, on time & 356 & $+480$ & 480 & 0.09 & 0.006 & 0.01 & 310\\
\midrule
recorded, human$\to$agent & 384 & $+240$ & 320 & 0.22 & 0.031 & 0.16 & 103\\
\bottomrule
\end{tabular}
\end{table}

\noindent\textbf{Per-conversation stability of the delay sweep.}
The paper's modes are pooled over conversations. Per conversation, the modal offset of the delayed direction minus the
injected delay has a median of 6, 6, 5.5, 5, and 4.5 frames at $D=0$, 3, 8,
10, 15; it lies within one frame of $6+D$ in 26/36, 28/35, 22/36, 25/35, and
18/36 conversations at those levels, and exactly at $6+D$ in 15, 14, 16, 14,
and 8 of them. Conversations with fewer than three transfers in that
direction are excluded, and when two frames tie for the mode the earlier one
is taken, a rule that leans against the anchoring claim. The delayed side's share of transfers that collide with
the partner's onset rises from 0.17 at $D=0$ to 0.36 at $D=15$ while the
on-time side's falls from 0.18 to 0.09 (Table~\ref{tab:rows}, overlap).

\noindent\textbf{Recorded partner.}
Each of 72 agent runs faces a 90\,s window of one AMI Headset-0 recording
(ES2003a or ES2004a; 24 windows at a 20\,s step), Mimi-encoded after
level-matching the recording to decoded Mimi speech (median of the loudest
15\% of 80\,ms frames scaled to 0.04 RMS; gains $\times7$--$\times14$), and
delivered one frame late for parity with the room's buffering. The human
side is segmented from its stored frame energy under the same 0.01 gate as
the agent side (voiced fraction median 0.31). Human$\to$agent transfers:
$n=384$, median 240\,ms, mode 4 frames holding 0.12 of transfers (frame 6 holds 0.08, against 0.23 in the matching direction of live dyads), overlap 0.22, band share 0.16.
A mismatched-window null that scores each agent's onsets against the window
a different agent heard (300 derangements) gives mode share 0.09
$[0.07,0.13]$ ($p=0.04$ for the observed 0.12) and band share 0.15
$[0.11,0.19]$. The agents' speech share is 0.15 against 0.35 in the
baseline dyads. Two earlier recordings (ES2002a, ES2005a) were excluded
before analysis because their crosstalk sits at the gate after
level-matching (more than a quarter of frames within a factor of two of
0.01 RMS).

\section{Clip, grid, and segmentation sensitivity}
\noindent\textbf{Clip.}
$W_1$ is computed between distributions clipped at $\pm2$\,s. The clip
holds 0.7\% of the human reference (0.05\% beyond $\pm4$\,s), 0.5\% of the
baseline energy transfers, 0.1\% of the forced-opener arm, 0.2\%, 0.1\%,
0.1\%, and 0.7\% of the $D=3$, 8, 10, 15 arms, 0.8\% of the recorded-partner
arm, and 38.3\% of the silence control's 26 transfers. Widening the clip to
$\pm4$\,s moves the baseline energy $W_1$ from 228 to 231\,ms.

\noindent\textbf{Grid.}
Agent offsets lie on the 80\,ms frame grid; human word boundaries are
continuous. Rounding the human boundaries outward to the grid (starts
floored, ends ceiled) moves the human median from 137 to 80\,ms and raises
the baseline energy $W_1$ from 228 to 266\,ms, so the grid works against
the deficits the paper reports, not for them. The paper keeps continuous
human boundaries.

\noindent\textbf{Backchannel-like material.}
No lexical backchannel filter is applied on either side. Under the shared
rule the share of different-speaker adjacencies that are nested (and
therefore not transfers) is 14.2\% in Switchboard (31{,}355 of 221{,}502)
and 16.3\%, 13.7\%, and 14.1\% for the agents under the energy, text-word,
and aligned-word segmentations; the share of IPUs shorter than 1\,s is
42.6\% for humans and 45.2\%, 84.3\%, and 50.1\% for the agents.

\noindent\textbf{Phase-matched reference.}
Restricting the reference to the first 90\,s of each Switchboard call gives
43{,}890 transfers, median 163\,ms, overlap 0.28, band share 0.25, and
lowers the baseline energy $W_1$ from 228 to 215\,ms. Shifting this
reference to each agent median gives the median-matched comparison:
overlap 0.175 against 0.084 (energy), 0.142 against 0.058 (text), 0.183
against 0.119 (aligned); band share 0.033 against 0.095, 0.088 against
0.072, and 0.043 against 0.122.

\section{Archived inventory}
Status is relative to $\alpha_*$: Below, Nominal ($\alpha_*<p\le0.05$), or
Above. Archive labels are kept for traceability; ``pp\_F'' is the baseline,
``pp\_G'' the forced-opener arm, ``pp\_deaf'' the silence control.
\small
\begin{longtable}{@{}r p{2.15cm} p{7.0cm} c r l@{}}
\caption{The 31 entries in the archived multiplicity inventory.}\label{tab:inventory}\\
\toprule
ID & Family & Contrast & Method & Probability & Status\\
\midrule\endfirsthead
\toprule ID & Family & Contrast & Method & Probability & Status\\\midrule\endhead
\bottomrule\endfoot
1 & T1 floor & energy: median $\ne$ human & CB & 0.0002 & Below \\
2 & T1 floor & energy: overlap $\ne$ human & CB & 0.0002 & Below \\
3 & T1 floor & energy: band share $\ne$ human & CB & 0.0002 & Below \\
4 & T1 floor & text words: median $\ne$ human & CB & 0.0002 & Below \\
5 & T1 floor & text words: overlap $\ne$ human & CB & 0.0002 & Below \\
6 & T1 floor & text words: band share $\ne$ human & CB & 0.0002 & Below \\
7 & T1 floor & aligned words: median $\ne$ human & CB & 0.0002 & Below \\
8 & T1 floor & aligned words: overlap $\ne$ human & CB & 0.0002 & Below \\
9 & T1 floor & aligned words: band share $\ne$ human & CB & 0.0002 & Below \\
10 & T2 re-paired & pp\_F: median vs surrogate & DP & 0.0005 & Below \\
11 & T2 re-paired & pp\_F: overlap vs surrogate & DP & 0.0005 & Below \\
12 & T2 re-paired & pp\_F: IQR vs surrogate & DP & 0.0005 & Below \\
13 & T2 re-paired & pp\_F: $W_1$ vs surrogate & DP & 0.0005 & Below \\
14 & T2 re-paired & pp\_G: median vs surrogate & DP & 0.0005 & Below \\
15 & T2 re-paired & pp\_G: overlap vs surrogate & DP & 0.0005 & Below \\
16 & T2 re-paired & pp\_G: IQR vs surrogate & DP & 0.0005 & Below \\
17 & T2 re-paired & pp\_G: $W_1$ vs surrogate & DP & 0.0005 & Below \\
18 & T2 re-paired & pp\_deaf: median vs surrogate & DP & 0.319 & Above \\
19 & T2 re-paired & pp\_deaf: overlap vs surrogate & DP & 0.775 & Above \\
20 & T2 re-paired & pp\_deaf: IQR vs surrogate & DP & 0.955 & Above \\
21 & T2 re-paired & pp\_deaf: $W_1$ vs surrogate & DP & 0.425 & Above \\
22 & T2 silence & 12/12 cells: more transfers with partner audio & EB & 0.00049 & Below \\
23 & T3 markers & forced, 0--10\,s & SF & 0.0015 & Nominal \\
24 & T3 markers & forced, 10--30\,s & SF & 0.11 & Above \\
25 & T3 markers & forced, 30--90\,s & SF & 0.16 & Above \\
26 & T3 markers & forced, 0--90\,s & SF & 0.0002 & Below \\
27 & T3 markers & forced, excl.\ tokens, 0--10\,s & SF & 0.060 & Above \\
28 & T3 markers & forced, excl.\ tokens, 0--90\,s & SF & 0.0002 & Below \\
29 & T3 role split & opener-holder, 0--90\,s & SF & $5\times10^{-5}$ & Below \\
30 & T3 role split & partner, 0--90\,s & SF & 0.12 & Above \\
31 & $W_1$ null & 36/36 conversations above their clustered null & EB & $2.9\times10^{-11}$ & Below \\
\end{longtable}
\normalsize

\section{Opening register, in full}
The lexicon has 21 patterns in three categories (greeting-by-name,
service offers, telephone framing) and was written after inspecting the
baseline transcripts; the baseline holds 58 markers, 38 of them in the
first 10\,s (66\% of markers in 14\% of the words). The five openers are
one scene-appropriate sentence per scene, placed in the first speaker's
speech with timing left to the agent. Table~\ref{tab:opener} gives every
window and contrast; the paper's Fig.~3b plots the 0--10, 10--30, and
30--90\,s rates of all three arms and its text quotes the 0--10 and
0--90\,s contrasts.

\begin{table}[ht]
\centering\small
\caption{Opening-register markers per 1k words, paired over 36 seeded
conversations. Base and treated are means of per-conversation rates over the
pairs with words in both arms, which is why they differ slightly from the
per-arm means of the paper's Fig.~3b in the later windows (28 pairs against
31 and 33 conversations at 30--90\,s); $\Delta$ is the mean paired
difference, with the sign-flip interval and two-sided $p$.}
\label{tab:opener}
\setlength{\tabcolsep}{3.5pt}
\begin{tabular}{@{}llrrrl@{}}
\toprule
contrast & window & base & treated & $\Delta$ [95\% CI] & $p$\\
\midrule
forced vs.\ baseline & 0--10\,s & 29.10 & 12.21 & $-16.9$ $[-27.7,-6.0]$ & 0.0015\\
 & 10--30\,s & 3.81 & 1.13 & $-2.7$ $[-5.7,+0.2]$ & 0.11\\
 & 30--90\,s & 2.31 & 0.98 & $-1.3$ $[-3.1,+0.4]$ & 0.16\\
 & 10--90\,s & 2.74 & 0.78 & $-2.0$ $[-3.5,-0.5]$ & 0.0082\\
 & 0--90\,s & 7.64 & 2.21 & $-5.4$ $[-8.8,-2.2]$ & 0.0002\\
forced, tokens excluded & 0--10\,s & 29.10 & 17.70 & $-11.4$ $[-23.3,+0.5]$ & 0.060\\
 & 0--90\,s & 7.64 & 2.31 & $-5.3$ $[-8.7,-2.1]$ & 0.0002\\
forced, opener-holder only & 0--90\,s & & & $-7.4$ $[-11.2,-3.7]$ & $5\times10^{-5}$\\
forced, partner only & 0--90\,s & & & $-3.5$ $[-7.9,+0.8]$ & 0.12\\
forced, dominant pattern removed & 0--90\,s & & & $-3.0$ & 0.009\\
prompt vs.\ baseline & 0--10\,s & 29.10 & 26.70 & $-2.4$ $[-14.3,+9.5]$ & 0.68\\
 & 0--90\,s & 7.64 & 5.09 & $-2.6$ $[-5.8,+0.6]$ & 0.12\\
 & 10--90\,s (34 pairs) & 2.81 & 1.19 & $-1.6$ $[-3.3,+0.0]$ & 0.055\\
prompt vs.\ forced & 0--90\,s & 2.21 & 5.09 & $+2.9$ $[+0.6,+5.2]$ & 0.012\\
\bottomrule
\end{tabular}
\end{table}

\noindent\textbf{Prompt compliance.}
The prompt condition gives the same sentence to the same agent as a
system-prompt instruction (``Your first words to them are: \ldots''), with
the same seeds. Compliance is the share of the opener's content words
(longer than three letters) that occur anywhere in the instructed agent's
first 60 words; 8 of 36 agents reach 80\%, 8 of 36 produce none. The
10--90\,s prompt contrast retains the 34 pairs with words in both arms.

\section{Within-conversation drift}
Human offsets tighten within a conversation at $-1.45$\,ms/s (SE 0.08);
the agents' slope is $-0.34$ (SE 0.69). A conversation fixed-effects model
with cluster-robust uncertainty gives a system$\times$time interaction of
$+1.11$\,ms/s, 95\% CI $[-0.24,+2.45]$, $p=0.11$ (wild cluster bootstrap
0.12); the design has 56\% power against a human-sized effect. The paper
does not report this contrast as a finding.

\section{The timing axis}
Let $t_s$ be the source speaker's IPU end, $t_r^{(q)}$ the time the
corresponding audio becomes readable at codebook $q$, and $t_o$ the
responding agent's onset, all on the shared clock. With base codebook delay
$\delta_q$ (160\,ms coarse, 240\,ms fine) and added one-sided delay $D$
frames,
\[
 t_r^{(q)}=t_s+\delta_q+80D\ \mathrm{ms},\qquad
 \mathrm{FTO}=t_o-t_s=\delta_q+80D\ \mathrm{ms}+(t_o-t_r^{(q)}).
\]
The paper's offsets are $t_o-t_s$. The observed modes of $6+D$ frames
leave a 480\,ms mode after subtracting $80D$; that residual still contains
$\delta_q$, so the wait after the fine path is about 240\,ms. The sweep
establishes that the wait is anchored to $t_r$ and is not shortened as $D$
grows; it does not by itself separate a fixed threshold from a projection
that starts after the perceived end. What separates them in the paper is
the distribution's shape at $D=0$: 0.011 of agent transfers fall in
$[-120,0)$\,ms against 0.104 for humans, and 90\% of positive agent
offsets exceed 320\,ms against 42\% for humans, so onsets in the approach
to the perceived end, which projection produces and endpointing cannot, are
all but absent; the agents' remaining overlaps lie deeper in the partner's turn (among
overlaps beyond 120\,ms, median $-560$\,ms against $-311$\,ms for the human
ones). Chang et al.'s generate-to-perceive lag of about
0.6\,s under interruption is consistent with the residual and is not
identified by floor timing alone.

\end{document}